\documentclass[conference]{IEEEtran}
\IEEEoverridecommandlockouts
\usepackage{cite}
\usepackage{amsmath,amssymb,amsfonts}
\usepackage{graphicx}
\usepackage{textcomp}
\usepackage{xcolor}
\def\BibTeX{{\rm B\kern-.05em{\sc i\kern-.025em b}\kern-.08em
    T\kern-.1667em\lower.7ex\hbox{E}\kern-.125emX}}

\usepackage{algorithm}
\usepackage[noend]{algorithmic}
\usepackage[font=footnotesize]{subcaption} 

\usepackage{hyperref}

\usepackage{amsmath}

\DeclareMathOperator{\bD}{\mathbf{D}}

\DeclareMathOperator{\bX}{\mathbf{X}}

\DeclareMathOperator{\bY}{\mathbf{Y}}
\DeclareMathOperator{\bA}{\mathbf{A}}

\DeclareMathOperator{\bW}{\mathbf{W}}

\DeclareMathOperator{\bR}{\mathbf{R}}
\DeclareMathOperator{\bP}{\mathbf{P}}

\begin{document}

\title{StitchNet: Composing Neural Networks from Pre-Trained Fragments}

% \author {
%     % Authors
%     Surat Teerapittayanon,
%     Marcus Comiter,
%     Brad McDanel,
%     H.T. Kung
% }
\author{\IEEEauthorblockN{Surat Teerapittayanon}
\IEEEauthorblockA{
\textit{National Nanotechnology Center}\\
Pathum Thani, Thailand \\
surat.tee@nanotec.or.th}
\and
\IEEEauthorblockN{Marcus Comiter}
\IEEEauthorblockA{
\textit{Harvard University}\\
Cambridge, MA, USA \\
marcuscomiter@post.harvard.edu}
\and
\IEEEauthorblockN{Bradley McDanel}
\IEEEauthorblockA{
\textit{Franklin \& Marshall College}\\
Lancaster, PA, USA \\
bmcdanel@fandm.edu}
\and
\IEEEauthorblockN{H.T. Kung}
\IEEEauthorblockA{
\textit{Harvard University}\\
Cambridge, MA, USA \\
kung@harvard.edu}
}

\maketitle

\begin{abstract}
We propose StitchNet, a novel neural network creation paradigm that stitches together fragments (one or more consecutive network layers) from multiple pre-trained neural networks. 
StitchNet allows the creation of high-performing neural networks without the large compute and data requirements needed under traditional model creation processes via backpropagation training. 
We leverage Centered Kernel Alignment (CKA) as a compatibility measure to efficiently guide the selection of these fragments in composing a network for a given task tailored to specific accuracy needs and computing resource constraints. 
We then show that these fragments can be stitched together to create neural networks with accuracy comparable to that of traditionally trained networks at a fraction of computing resource and data requirements.
Finally, we explore a novel on-the-fly personalized model creation and inference application enabled by this new paradigm.
The code is available at \href{https://github.com/steerapi/stitchnet}{https://github.com/steerapi/stitchnet}.
\end{abstract}


\section{Introduction}
AI models have become increasingly more complex to support additional functionality, multiple modalities, and higher accuracy. While the increased complexity has improved model utility and performance, it has imposed significant model training costs. Therefore, training complex models is often infeasible for resource-limited environments, such as those at the cloud edge and in fully or often disconnected environments. 

\begin{figure}[t!]
    \centering
    \includegraphics[width=0.9\linewidth]{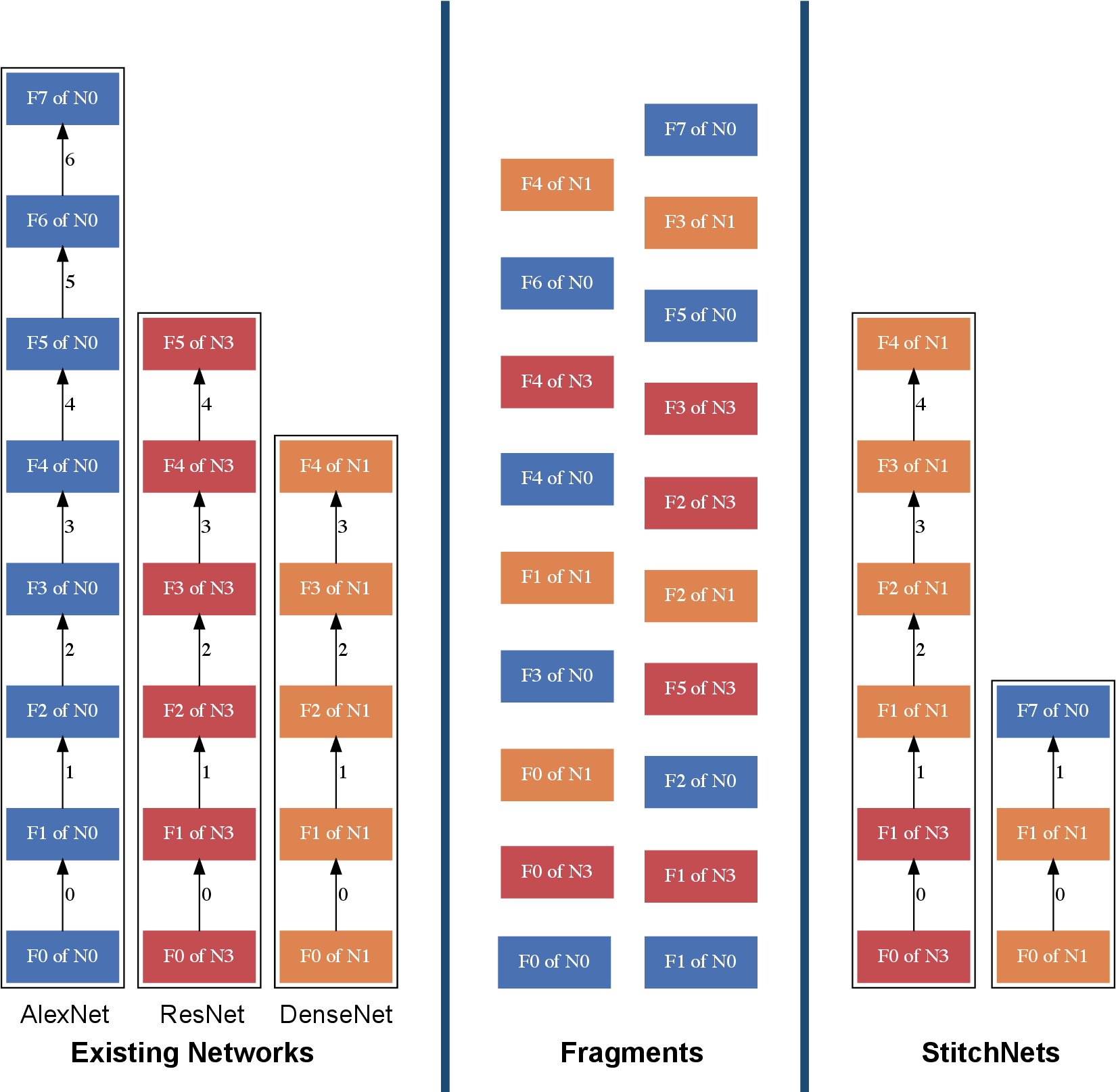}
    \caption{Overview of the StitchNet approach. 
    Existing networks (left) are cut into fragments (middle), which are composed into StitchNets (right) created for a particular task. No retraining is needed in this process.}
    \label{fig:overview}
\end{figure}

In response to these challenges, this paper proposes a new paradigm for creating neural networks: rather than training networks from scratch or retraining existing networks, we create neural networks through composition by \textit{stitching together} fragments of existing pre-trained neural networks. A fragment is one or more consecutive layers of a neural network. We call the resulting neural network composed of one or more fragments a ``StitchNet'' (Fig. \ref{fig:overview}). By significantly reducing the amount of computation and data resources needed to create neural networks, StitchNet enables an entire new set of applications, such as the rapid generation of personalized neural networks at the edge or in fully disconnected environments.

StitchNet's model creation mechanism is fundamentally different from today's predominant backpropagation-based method for creating neural networks. Given a dataset and a task as input, the traditional training method uses backpropagation with stochastic gradient descent (SGD) or other optimization algorithms to adjust the weights of the network. This training process iterates through the full dataset multiple times, and therefore requires compute resources that scale with the amount of data and the complexity of the network. Training large models in this way also requires substantial amounts of data to mitigate overfitting.  While successful, this traditional paradigm for model creation is not without its limitations, especially as AI moves out of the data center and into highly resource-constrained and disconnected environments.  Creating complex neural networks without access to large amounts of data and compute resources is a growing challenge. In the extreme case (e.g., for very large language models (LLMs) and computer vision models), only a few companies with access to unrivaled amounts of data and compute resources are able to create such models.

StitchNet solves this problem by creating new neural networks using fragments of already existing neural networks. The new approach takes advantage of the growing amount of neural networks that already exist. %, having been trained previously by many groups and companies. 
% \textbf{Just as large language models (LLMs) take advantage of the large amount of written content in existence, StitchNet is a method for taking advantage of the large amount of existing neural networks in existence.}
 StitchNet enables the efficient reuse of the learned knowledge resident in those pre-trained networks, which has been distilled from large amounts of data, rather than having to relearn it over and over again for new tasks as is done with traditional model creation paradigms.  StitchNet's ability to reuse existing pre-trained fragments, rather than recreating from scratch or retraining for every task, will help accelerate the growth and application of neural networks for solving more and more complex tasks in heterogenous environments.  For example, StitchNets can be created on-the-fly to serve as classifiers specially tuned for local conditions and classes of interest in a given environment or task.

However, compositing these existing fragments into a coherent and high-performing neural network is non-trivial. To reuse the knowledge of pre-trained neural network fragments, we need a way to 1) measure the compatibility between any two fragments, and 2) compose compatible fragments together.  In the past, Centered Kernel Alignment (CKA)~\cite{kornblith2019similarity, cortes2012algorithms, cristianini2006kernel} has been used to measure similarity between neural network representations. We leverage CKA to assess the compatibility of any two fragments from any neural networks, and compose new neural networks from fragments of existing pre-trained neural networks to create high-performing networks customized for specific tasks without the costs of traditional model creation methods. The CKA score is used to reduce the search space to identify compatible fragments and guide the fragment selection process. 

We present empirical validations on benchmark datasets, comparing the performance of StitchNet to that of the original pre-trained neural networks.  We demonstrate that StitchNet achieves comparable or higher accuracy on personalized tasks compared with off-the-shelf networks and has significantly lower computational and data requirements than training networks from scratch or by fine-tuning.
 
Our contributions are:
\begin{itemize}
\item The StitchNet paradigm: a novel neural network creation method with versatile applications.
\item Innovative use of CKA to assess fragment compatibility.
\item Technique for seamlessly combining compatible fragments in both linear and convolutional layers.

% \item The StitchNet paradigm: We introduce a novel neural network creation method called StitchNet, which opens up new possibilities for a wide range of applications.

% \item Centered Kernel Alignment (CKA) application: We propose a novel use of CKA to assess the compatibility of any two fragments for their composition. This approach enhances the understanding of fragment compatibility in neural network design.

% \item Composing compatible fragments: We present a technique for effectively combining compatible fragments, enabling seamless integration of both linear and convolutional layers. This method ensures efficient and optimized neural network composition.

% \item The StitchNet paradigm: a novel neural network creation method that enables a new set of applications.
% \item A novel use of Centered Kernel Ailgnment (CKA) in assessing the compatibility of any two fragments for their composition.
% \item A technique for composing compatible fragments together for both linear and convolutional layers.
% \item A feasibility demonstration of StitchNets for efficient on-the-fly personalized neural network creation and inference.
\end{itemize}

\section{Composing Fragments}

The core mechanism to create StitchNets is to identify reusable fragments from a pool of existing networks and to compose them into a coherent neural network model capable of performing a given task.  To this end, we need a way to determine how compatible any two candidate fragments are with each other. In previous work, Kornblith, et al.~\cite{kornblith2019similarity} presented centered kernel alignment (CKA) \cite{cortes2012algorithms, cristianini2006kernel} as a way to measure the similarity between neural network representations. Rather than looking at the neural network as a whole, we adopt and use CKA to as a measure of compatibility between any two \textit{fragments} of any neural networks. 

In this section, we first define CKA as a way to measure how compatible any two fragments are with one another and therefore their ability to be composed. Using CKA, we then present a technique to stitch different fragments together. Finally, we describe the algorithm to generate StitchNets.

\subsection{Centered Kernel Alignment (CKA)}
\label{section:score}
We define $\bX \in \mathbb{R}^{p \times n}$ as outputs of a fragment $F_A$ of model $A$ and $\bY \in \mathbb{R}^{q \times n}$ as inputs of a fragment $F_B$ of model $B$ of the same dataset $\bD$, where $n$ is the number of samples in the dataset, $p$ is the output dimension of $F_A$, and $q$ is the input dimension of $F_B$. Given a target dataset $\bD$, we define the compatibility score between any two fragments as $\text{CKA}(\bX, \bY)$ (specifically linear CKA) of fragment $F_A$ and fragment $F_B$:
\begin{equation}
\label{eq:cka}
\frac{\| \text{cov}(\bX^T\bX,\bY^T\bY) \|_F^2}{\sqrt{\| \text{cov}(\bX^T\bX,\bX^T\bX) \|_F^2\| \text{cov}(\bY^T\bY,\bY^T\bY) \|_F^2}}.
\end{equation}
To reduce memory usage for a large $\bD$, CKA can be approximated by averaging over minibatches as presented in \cite{nguyen2020wide}.

\subsection{Stitching Fragments}
\label{section:stitch}

Once we have determined compatible fragments based on a configurable compatibility threshold, the next step in creating a StitchNet is to \textit{stitch} the two fragments together. To do so, we find a projection tensor $\bA$ that projects the output space of one fragment to the input space of the other fragment we are composing.  We now describe this.

Without loss of generality, we assume the output and input tensors are 2D tensors, where the first dimension is the sample dimension. If the tensors are not 2D tensors, we first flatten all other dimensions with the exception of the sample dimension.

We use Einstein summation notation, where $i$ represents the sample dimension, $j$ the output dimension of the incoming fragment, and $k$ the input dimension of the outgoing fragment. Given an output tensor $\bX_{ij}$ of the incoming fragment and an input tensor $\bY_{ik}$ of the outgoing fragment, we seek to find $\bA$ such that
$\bY_{ik} = \bA_{kj}\bX_{ij}.$ We can then solve for $\bA$ using the Moore-Penrose pseudoinverse:
\begin{equation}
\label{eq:A}
    \bA_{kj} = \bY_{ik}\bX_{ij}^T(\bX_{ij}\bX_{ij}^T).^{-1}
\end{equation}

Once $\bA$ is found, we fuse $\bA$ with the weight of the first layer of the outgoing fragment. For linear layers, we simply do the following: $\bW_{lk}' = \bW_{lj}\bA_{kj},$
% \begin{equation}
% \label{eq:W}
%     \bW_{lk}' = \bW_{lj}\bA_{kj},
% \end{equation}
where $l$ is the dimension of the output feature of the outgoing layer.

For convolutional layers, we first upsample or downsample the spatial dimension to match each other, and then adjust the weight along the input channel dimension as follows: ${\bW_{okmn}'} = {\bW_{ojmn}}{\bA_{kj}},$
% \begin{equation}
% \label{eq:Wc}
%     {\bW_{okmn}'} = {\bW_{ojmn}}{\bA_{kj}},
% \end{equation}
where $o$ is the output channel dimension, $j$ is the original input channel dimension, $k$ is the new input channel dimension, and $m$ and $n$ are the spatial dimensions.

For stitching a convolutional layer with an output tensor $\bX$ and a linear layer with an input tensor $\bY$, we first apply adaptive average pooling so that the spatial dimension is 1x1 and flatten $\bX$ into a 2D tensor. Then, we follow Eq.~\ref{eq:A} to find $\bA$ and fuse it with the $\bW$ of the linear layer.

\subsection{StitchNet Generation}
\label{subsec:stitchnetgeneration}

We now describe the main algorithm for creating StitchNet networks (``StitchNet'' for short), shown in Algorithm~\ref{alg:stitchnet}. A StitchNet network is created by joining a set of pre-trained network fragments drawn from a pool $\bP = \{F_{ijk}\}$. We use the notation $F_{ijk}$ to denote a fragment of a neural network $i$ from layer $j$ to layer $k$, and the notation $N_{ik}$ to denote the computation performed by the portion of the neural network from which the fragment was taken up to layer $k$.  Other than the fragment pool $\bP$ and creation process hyperparameters ($K,T,L$), the only other input to the StitchNet creation process is a dataset $\bD$ for which the StitchNet will be optimized.  The dataset $\bD$ can be chosen to include relevant samples and classes of interest for the specific task at hand, as illustrated in Sec. ~\ref{subsec:dataset},~\ref{subsec:differenttasks}, and~\ref{subsec:onthefly}.
\begin{algorithm}[H]
\begin{algorithmic}[1]
   \STATE {\bfseries Input:} fragment pool $\bP = \{F_{ijk}\}$, network ${i}$ in $\bP$ up to layer $k$ $N_{ik}$, fragment starting with layer $j$ and ending in layer $k$ of network $i$ $F_{ijk}$, target dataset $\bD$ with $M$ samples, span $K$, threshold $T$, maximum number of fragments $L$, result set of StitchNets and their associated scores $\bR$, current StitchNet $Q$, current score $s$
   \STATE {\bfseries Output:} result set of StitchNets and their associated scores $\bR$
   \IF{$Q$ is empty}
      \STATE $\{F_{ijk}\}$ = select starting fragments in $\bP$
      \FOR{$F_{ijk}$ in $\{F_{ijk}\}$}
        \STATE StitchNet($\bP$, $\bD$, $K$, $T$, $L$, $\bR$, $F_{ijk}$, 1)
      \ENDFOR
   \ENDIF
   \IF{the number of fragments in $Q \ge L$}
      \RETURN $\bR$
   \ENDIF
   \STATE $\{F_{ijk}\}$ = select $K$ middle or terminating fragments in $\bP$
   \FOR{$F_{ijk}$ in $\{F_{ijk}\}$}
        \STATE $\bX$ = $Q$($\bD$); $\bY$ = $N_{ik}$($\bD$)
        \STATE $s_n$ = $s \times$ CKA$(\bX,\bY)$ (see section \ref{section:score})
        \IF{$s_n > T$}
            \STATE $Q$ = Stitch($Q$, $F_{ijk}$, $\bX$, $\bY$) (see section \ref{section:stitch})
            \IF{$F_{ijk}$ is a terminating fragment}
                \STATE $\bR$ = $\bR$ $\cup$ $\{Q,s_n\}$
            \ELSE
                \STATE StitchNet($\bP$, $\bD$, $K$, $T$, $L$, $\bR$, $Q$, $s_n$)
            \ENDIF
        \ENDIF
   \ENDFOR
   \RETURN $\bR$
\end{algorithmic}
   \caption{StitchNet($\bP$, $\bD$, $K$, $T$, $L$, $\bR$, $Q$, $s$)}
   \label{alg:stitchnet}
\end{algorithm}
We now describe the creation of the pool of network fragments $\bP$ derived from a set of pre-trained off-the-shelf networks.  These pre-trained networks are divided into one of three types of fragments: \textit{starting} fragments for which the input is the original network input, \textit{terminating} fragments for which the output is the original network output, and \textit{middle} fragments that are neither starting nor terminating fragments.

The first step in the StitchNet creation process is to choose the set of starting fragments. This could include all starting fragments in $\bP$, or a subset based on certain criteria, e.g., the smallest, biggest or closest starting fragment.

Once a set of starting fragments is selected, a StitchNet is built on top of each starting fragment that has a current starting score of 1. First, a set of $K$ candidate fragments are selected from $\bP$. These fragments can be selected based on CKA scores (i.e., $K$ fragments with the highest CKA scores), the number of parameters of the fragments (i.e., $K$ fragments with the least number of parameters in $\bP$), the closest fragments (i.e., $K$ fragments with the least latency in $\bP$ in a distributed setting), or other selection methods.

For each of the candidate fragments, we then compute two intermediate neural network computations in order to derive the compatibility score for further stitching.  First, we pass the dataset $\bD$ through the candidate StitchNet in its current form, resulting in the value $\bX$.  Second, we pass the same dataset $\bD$ through the neural network from which the candidate fragment $F_{ijk}$ was selected, resulting in the value $\bY=N_{ik}(D)$.  

After running these computations, we compute $\text{CKA}(\bX,\bY)$ as in Sec.~\ref{section:score}. We then multiply the CKA with the current score $s$ to obtain the new current score $s_n$. If $s_n$ is still greater than a set threshold $T$, the candidate fragment is selected and the process continues recursively. Otherwise, the candidate fragment is rejected.  The threshold can be set to balance the amount of exploration allowed per available compute resources.

This process continues until a terminating fragment is selected, the maximum number of fragments $L$ is reached, or all recursive paths are exhausted. At this point, the completed StitchNets and their associated scores $\bR$ are returned for user selection.

\section{Results}

We now demonstrate that StitchNets can perform comparably with traditionally trained networks but with significantly reduced computational and data requirements at both inference and creation time.
Through these characteristics, StitchNet enables the immediate on-the-fly creation of neural networks for personalized tasks without traditional training.  

\subsection{Fragment pool}
\label{subsec:fragmentpool}

To form the fragment pool $\bP$, we take five off-the-shelf networks pre-trained on the ImageNet-1K dataset~\cite{deng2009imagenet} from Torchvision~\cite{marcel2010torchvision}: \textit{alexnet}, \textit{densenet121}, \textit{mobilenet\_v3\_small}, \textit{resnet50} and \textit{vgg16} with IMAGENET1K\_V1 weights.\footnote{In practice, many more or specialized pre-trained networks can be included in the pool, including smaller and larger networks.}

\begin{figure}[th!]
    \centering
    \includegraphics[width=0.5\linewidth]{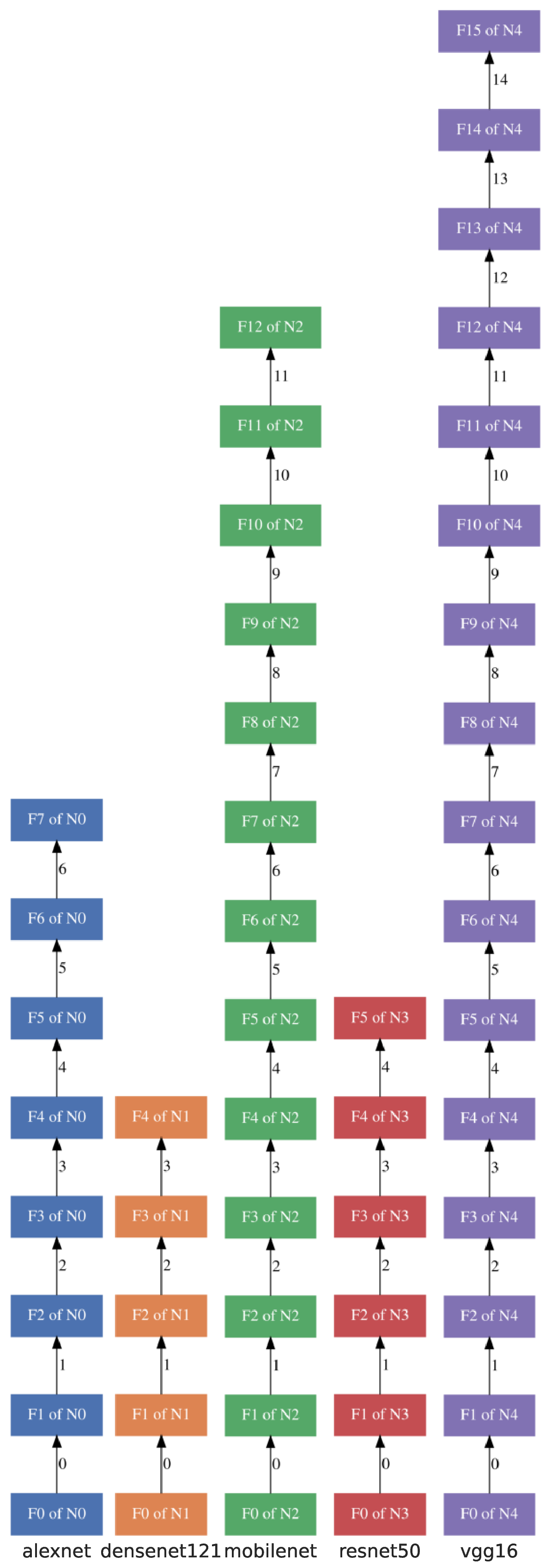}
    \caption{Five pre-trained networks are fragmented into a fragment pool $\bP$. These fragments will be stitched together to form StitchNets.}
    \label{fig:originalnets}
\end{figure}

These pre-trained networks are cut into fragments at each convolution and linear layer that has a single input. As shown in Fig.~\ref{fig:originalnets}, there are 8 fragments for \textit{alexnet}, 5 fragments for \textit{densenet121}, 13 fragments for \textit{mobilenet\_v3\_small}, 6 fragments for \textit{resnet50} and 16 fragments for \textit{vgg16}. This results in the creation of a fragment pool $\bP$ of 48 fragments consisting of 5 starting fragments, 38 middle fragments, and 5 terminating fragments. We use this fragment pool in all experiments. %in this paper.

\subsection{Dataset}
\label{subsec:dataset}

The dataset used to evaluate StitchNet in the first batch of experiments is the ``Dogs vs. Cats'' dataset~\cite{kaggle}. This dataset includes 25,000 training images of dogs and cats, and we use an 80:20 train:test split. 
We map ImageNet-1K class labels into cat and dog labels (class IDs 281-285 and 151-250, respectively). To form the target dataset $\bD$ for use in the stitching process of Algorithm~\ref{alg:stitchnet}, we randomly select $M$ samples from the training set as the target dataset $\bD$. We choose this task as it is characteristic of the type of task where StitchNet would be used: a user needs a custom classifier for a particular task and desired set of classes, but any general neural network problem formulation can also be used.

\subsection{StitchNet Generation}
\label{subsec:stitchnets}

We generate StitchNets with Algorithm~\ref{alg:stitchnet} using the fragment pool and the dataset described in Sec.~\ref{subsec:fragmentpool}~and~\ref{subsec:dataset}. We set $K=2$, $T=0.5$ and $L=16$. The number of samples $M$ in $\bD$ used for the stitching process is 32.

\begin{figure}[th!]
     \centering
     \begin{minipage}[t]{0.49\linewidth}
        %  \centering
        \includegraphics[width=\linewidth]{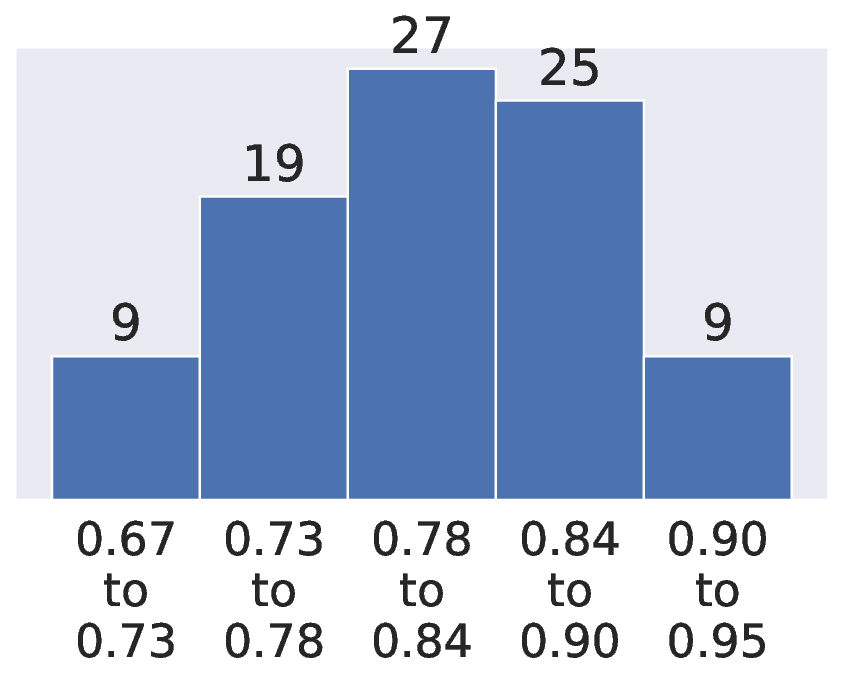}
        \subcaption{accuracy}
        \label{fig:hist_accuracy}
     \end{minipage}
     \hfill
     \begin{minipage}[t]{0.49\linewidth}
        %  \centering
        \includegraphics[width=\linewidth]{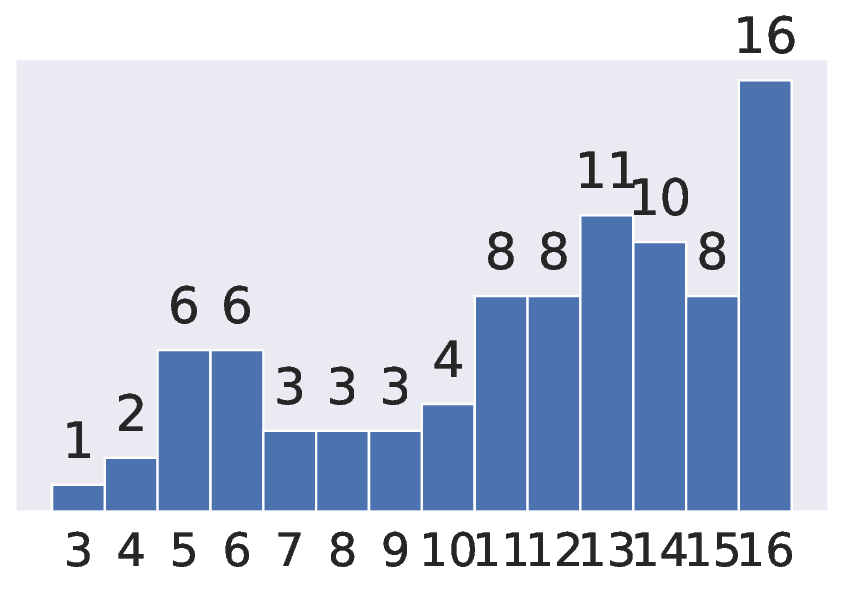}
        \subcaption{\# fragments}
        \label{fig:hist_fragments}
     \end{minipage}
      \vfill
     \begin{minipage}[t]{0.49\linewidth}
        %  \centering
        \includegraphics[width=\linewidth]{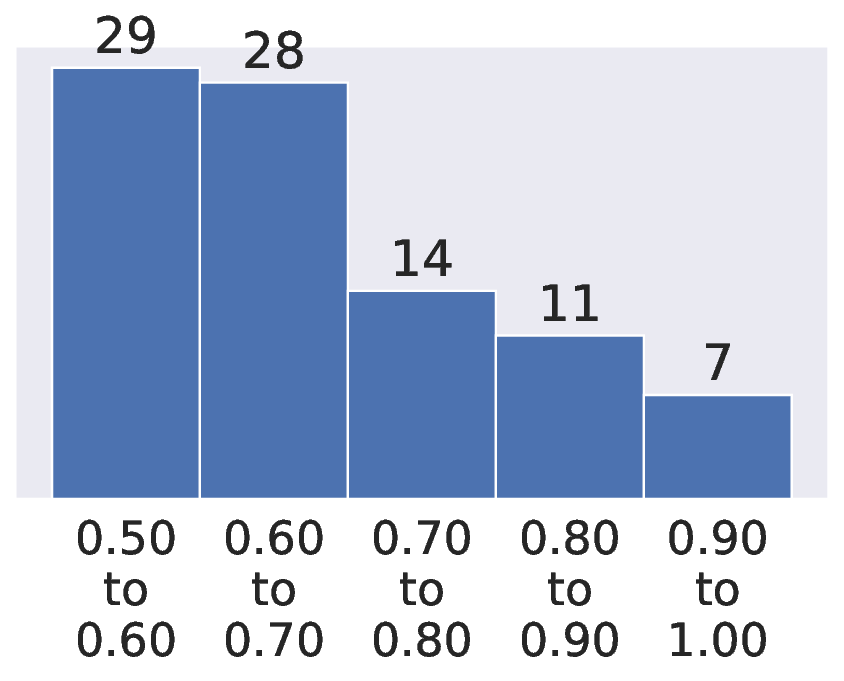}
        \subcaption{CKA score}
        \label{fig:hist_cka}
     \end{minipage}
     \hfill
     \begin{minipage}[t]{0.49\linewidth}
        %  \centering
        \includegraphics[width=\linewidth]{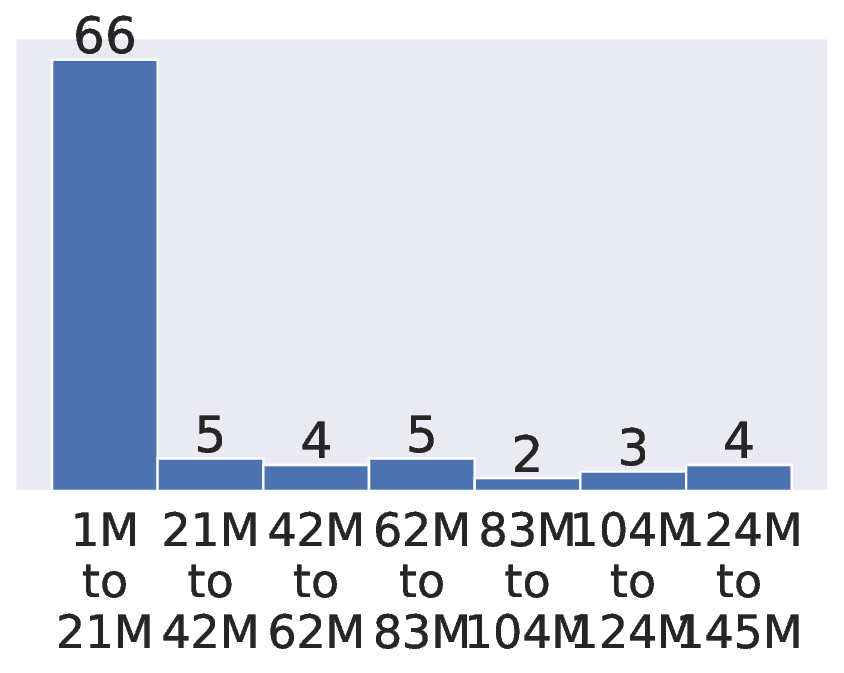}
        \subcaption{\# parameters}
        \label{fig:hist_parameters}
     \end{minipage}
    \caption{Histogram of (a) accuracy, (b) \# fragments, (c) CKA score, (d) \# parameters in the generated batch of StitchNets.}
    \label{fig:summarystatistics}
\end{figure}

Given these hyperparameters, a total of 89 StitchNets are generated.  We evaluate them on the test set of completely unseen test samples.  Summary statistics for the generated StitchNets are shown in Fig.~\ref{fig:summarystatistics}, including accuracy (\ref{fig:hist_accuracy}), number of fragments per StitchNet (\ref{fig:hist_fragments}), CKA score (\ref{fig:hist_cka}), and number of parameters per StitchNet (\ref{fig:hist_parameters}). 

\subsection{Reduction in Inference Computation}

We now demonstrate how StitchNet significantly reduces inference-time computational requirements over traditional neural network training paradigms by studying StitchNet accuracy as a function of parameters. 

\begin{figure}[th!]
     \centering
     \begin{minipage}[t]{\linewidth}
        %  \centering
        \includegraphics[width=\linewidth]{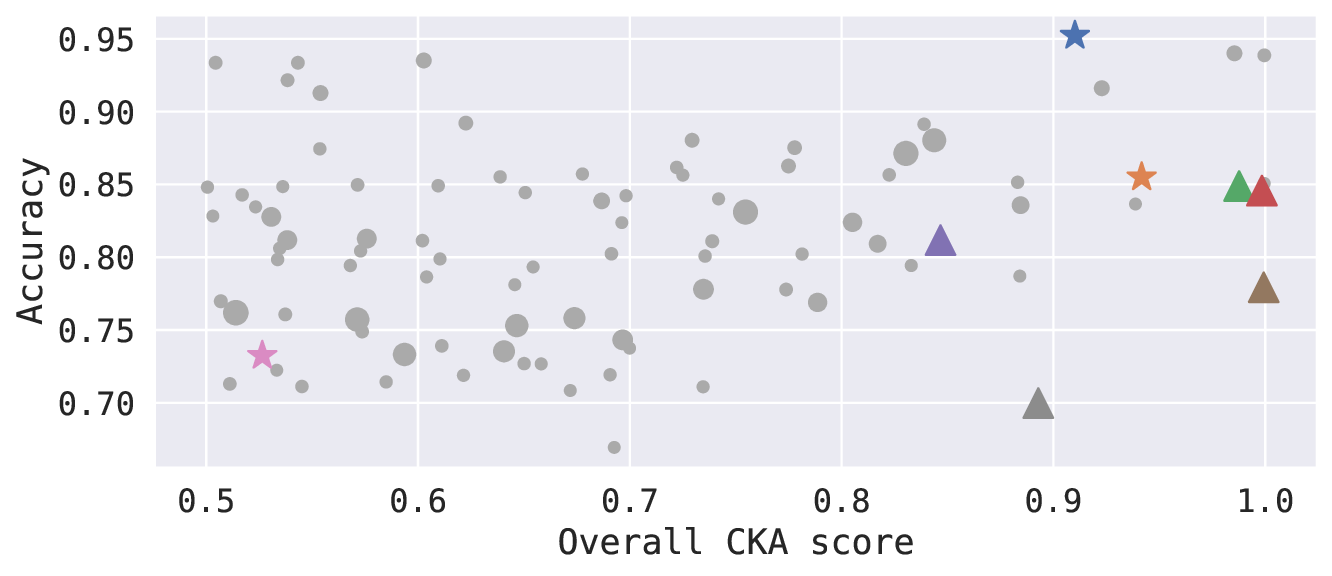}
        % \subcaption{Accuracy vs CKA}
        % \label{fig:accvscka}
     \end{minipage}
     % \vfill
     \begin{minipage}[t]{\linewidth}
        \vspace{-12px}
        %  \centering
        \includegraphics[width=\linewidth]{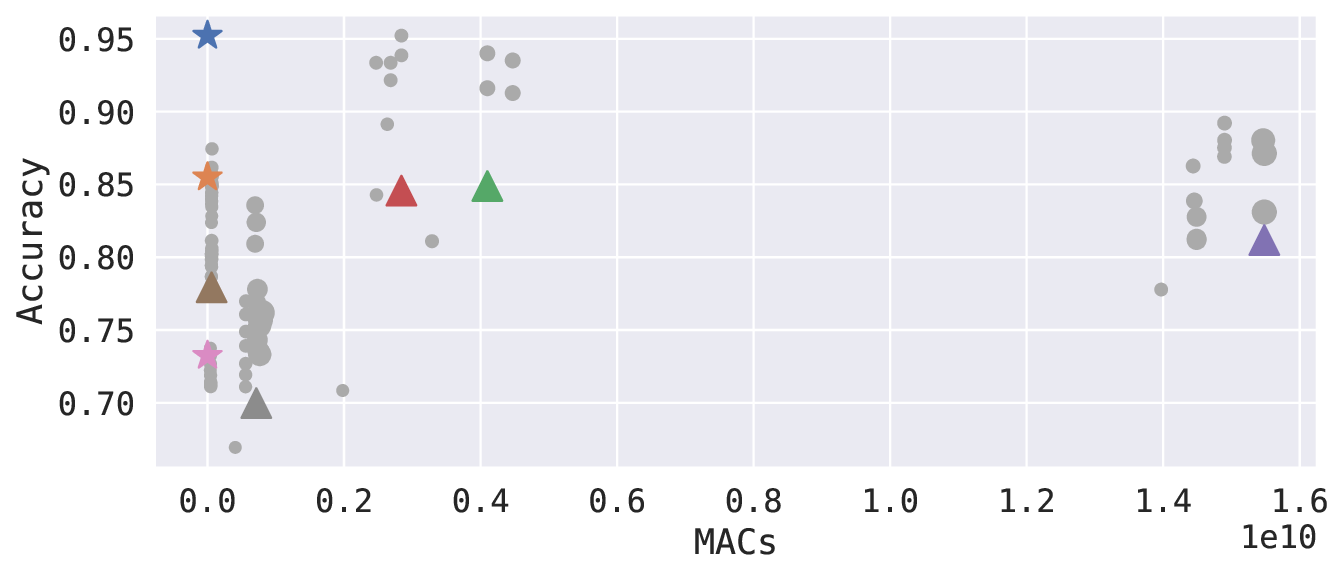}
        % \subcaption{Accuracy vs MACs}
        % \label{fig:accvsmacs}
     \end{minipage}
     % \vfill
     \begin{minipage}[t]{\linewidth}
        \vspace{-1px}
         \centering
        \includegraphics[width=.8\linewidth]{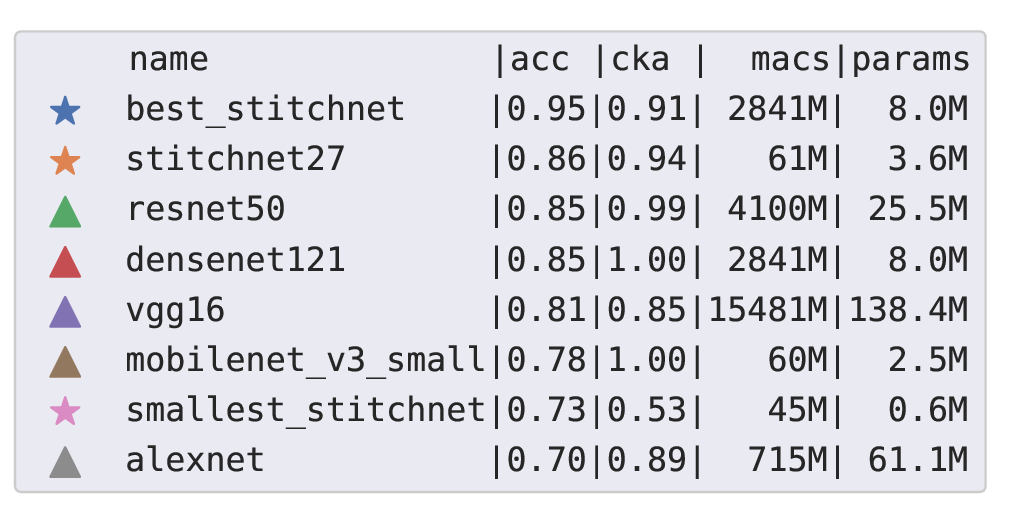}
        % \subcaption{Accuracy vs MACs}
        % \label{fig:accvsmacs_legend}
     \end{minipage}
    \caption{
    Accuracy vs Overall CKA score (top) and MACs (middle) on ``Cat vs. Dogs.'' $macs$ is the number of Multiply-Accumulate Operations (MACs), $acc$ is the accuracy and $cka$ is the overall CKA score. The gray dots are the StitchNets generated. The best StitchNet ($acc$=0.95) has 10-point higher accuracy than the best pre-trained model(s) (densenet121 and resnet50 with $acc$=0.85).
    }
    \label{fig:accvsckamacs}
\end{figure}

% \begin{figure*}[th!]
%     \centering
%     \includegraphics[width=\linewidth]{figures2/accvscka.eps}
%     \caption{
%     Accuracy vs the overall CKA score on ``Cat vs. Dogs.'' $cka$ is the overall CKA score, $acc$ is the accuracy and $macs$ is the number of Multiply-Accumulate Operations (MACs). The best StitchNet ($acc$=0.95) has 10\% higher accuracy than the best pre-trained model(s) (densenet121 and resnet50 with $acc$=0.85).
%     }
%     \label{fig:accvscka}
% \end{figure*}

% \begin{figure*}[th!]
%     \centering
%     \includegraphics[width=\linewidth]{figures2/accvsmacs.eps}
%     \caption{
%     Accuracy vs MACs on ``Cat vs. Dogs.'' $macs$ is the number of Multiply-Accumulate Operations (MACs), $acc$ is the accuracy and $cka$ is the overall CKA score. The best StitchNet ($acc$=0.95) has 10\% higher accuracy than the best pre-trained model(s) (densenet121 and resnet50 with $acc$=0.85).
%     }
%     \label{fig:accvsmacs}
% \end{figure*}

\begin{figure}[th!]
    \centering
    \includegraphics[width=0.7\linewidth]{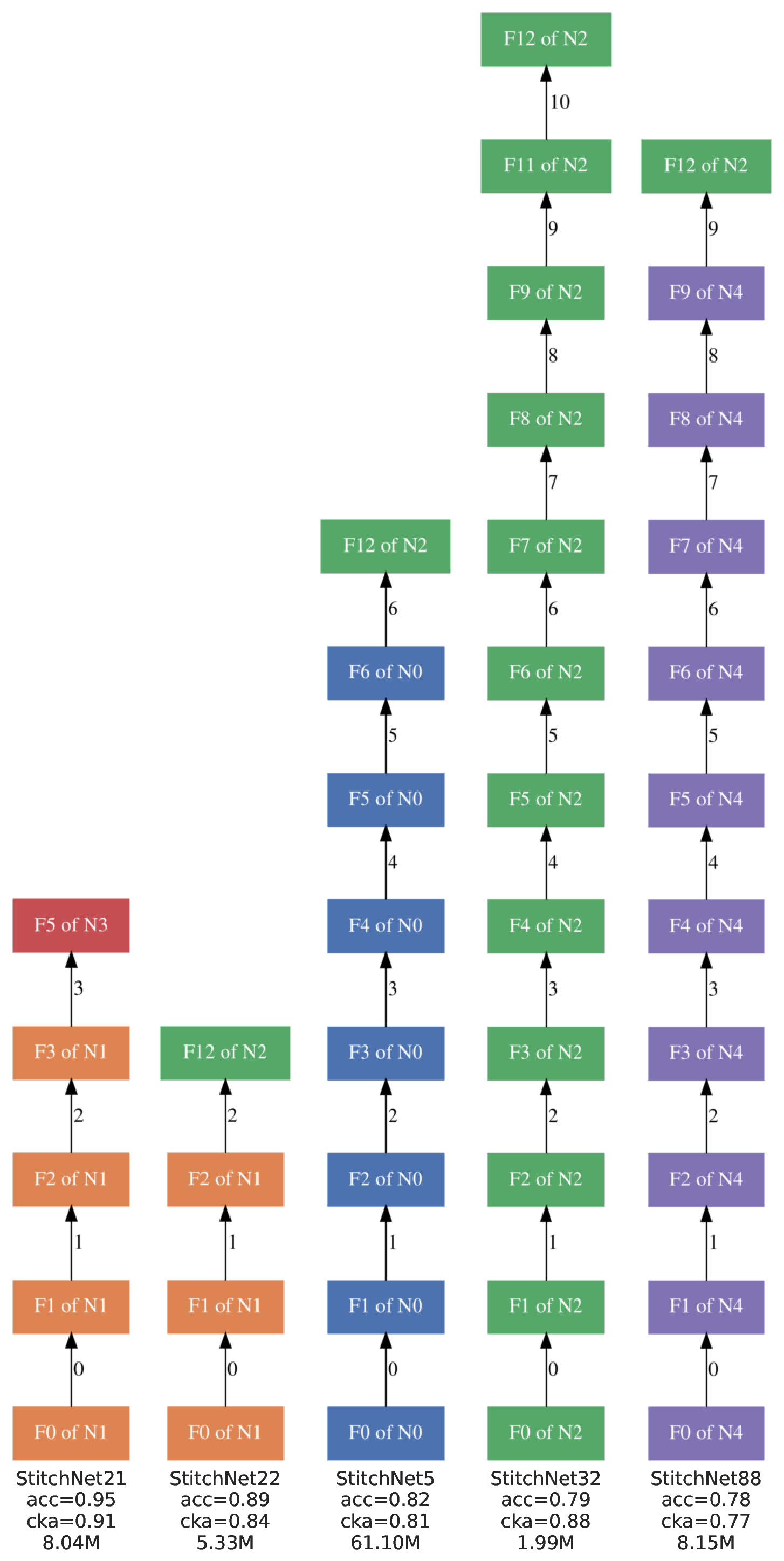}
    \caption{Examples of generated StitchNets.}
    \label{fig:stitchnets}
\end{figure}

Fig.~\ref{fig:accvsckamacs} shows the resulting accuracy of the generated StitchNets as a function of the overall CKA score (top) and Multiply-Accumulate Operations (MACs) (middle) for each generated StitchNet and the number of parameters (proportional to marker size) as a proxy for inference-time computation cost. We find a number of StitchNets outperform the pre-trained network while realizing significant computational savings.  For example, StitchNet27 (denoted by an orange star) achieves an accuracy of 0.86 with 3.6M parameters compared with the 0.70 accuracy of the pre-trained alexnet with 61.1M parameters. Therefore, StitchNet achieves a 16-point absolute increase in accuracy with a 94.1\% reduction in number of parameters for the given task when compared with those of the pre-trained alexnet.  Similar trends hold in comparison of StitchNets with the other pre-trained models.  Fig.~\ref{fig:stitchnets} shows the composition of some of these high-performing StitchNets, demonstrating the diversity in fragment use, ordering, and architectures.  

These results crystallize one of the core benefits of StitchNet: without significant data and computation requirements of traditional training procedures, the method can discover networks that are personalized for the task, outperform the original pre-trained networks, and do so while significantly reducing inference-time compute requirements. This is due to the fact that these pre-trained networks are not trained to focus on these two specific classes, while our StitchNets are stitched together specifically for the task.  In the next section, we will see that very little data is required for the stitching process.

% Additionally, we compare the StitchNets with the various off-the-shelf models, denoted by  triangles.  We find that the StitchNet generation process creates many different StitchNets that outperform the off-the-shelf models, many of which do so at reduced computational cost. Fig.~\ref{fig:stitchnets} shows the composition of some of these high-performing StitchNets, demonstrating the diversity in fragment use, ordering, and architectures. 

We also validate the effectiveness of using CKA to guide the stitching procedure.  StitchNets with high CKA scores (especially above 0.9) also have high accuracy. Therefore CKA can be a proxy to measure compatibility between connecting fragments.\footnote{Note that there exist high accuracy StitchNets with low overall CKA scores. This is because neural networks are highly redundant and invariant to small changes, making them able to tolerate a certain amount of errors while still providing quality predictions (see Sec.~\ref{section:discuss_why}).}  But while high CKA scores imply high accuracy, having high accuracy does not necessarily imply equally high CKA scores.  Regardless, CKA is still useful as a heuristic because it removes highly incompatible choices (i.e., those with very low CKA scores), therefore saving computation time in the stitching process.  These savings are substantial as there are more candidates with low CKAs than high CKAs.

\subsection{Reduction in Network Creation Computation}
\label{subsec:computerequirements}
We now demonstrate that StitchNet can be created without significant data and computation requirements. 
Specifically, we compare StitchNet21 (generated in Fig.~\ref{fig:stitchnets} on the target dataset of $M=32$ samples) with fine-tuning the same five off-the-shelf networks (retraining them using the training portion of dataset of Sec.~\ref{subsec:dataset}). For fine-tuning, we replace and train only the last layer of the pre-trained network using SGD with batch size 32, learning rate $0.001$ and momentum $0.9$. The results shown are averaged over 10 runs. For ease of comparison, we normalize the computation cost in terms of the number of samples processed through a neural network. In practice, fine-tuning requires backpropagation, which incurs additional computation per sample than StitchNet generation.

\begin{figure}[th!]
    \centering
    \includegraphics[width=\linewidth]{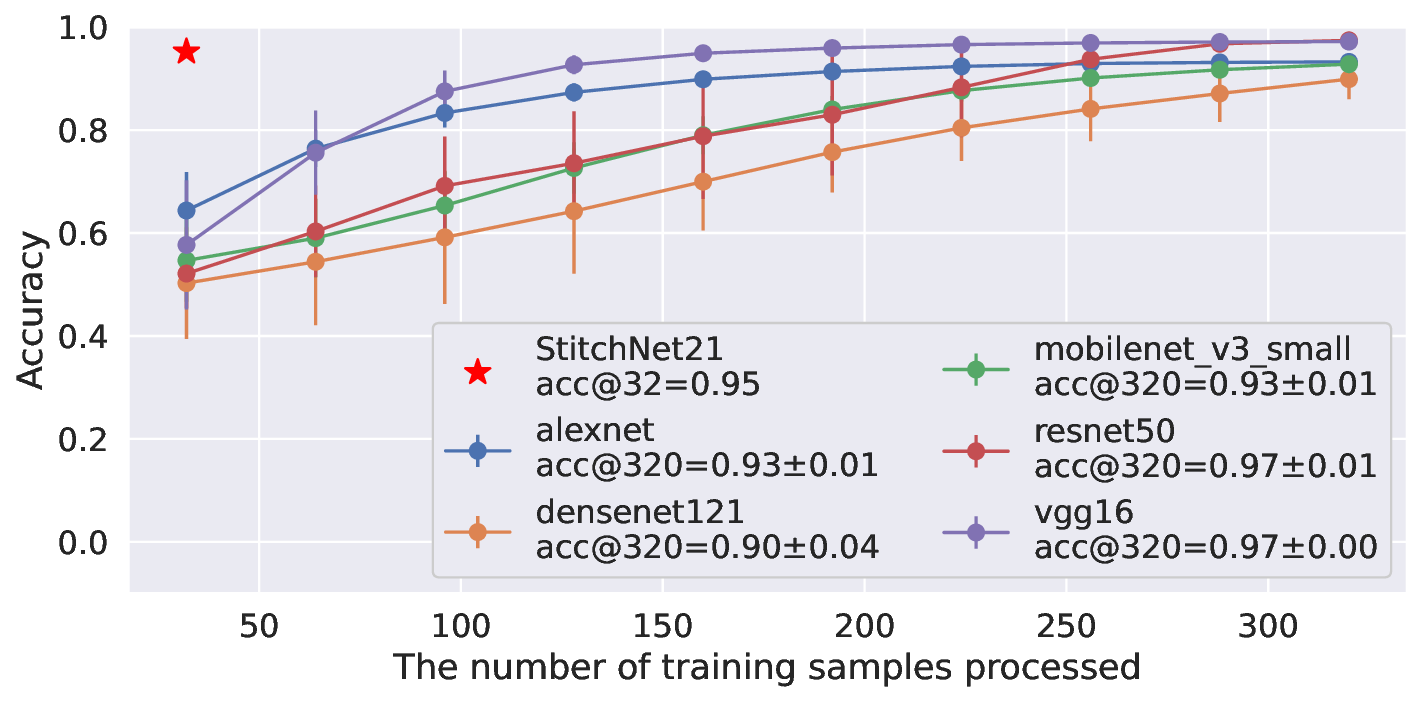}
    \caption{Accuracy vs the number of training samples processed (i.e., data and computation required). StitchNet requires only a fraction of the computation of traditional training methods to achieve comparable performance.}
    \label{fig:acc_data}
\end{figure}

Fig.~\ref{fig:acc_data} compares the accuracy of StitchNet21 (denoted by the red star) with the traditionally fine-tuned networks as a function of the number of training samples processed. For a given accuracy target, StitchNet processes a substantially smaller number of data samples than traditionally fine-tuned networks. Specifically, to reach an accuracy of 0.95, fine-tuning of \textit{alexnet}, \textit{densenet121}, and \textit{mobilenet\_v3\_small} requires processing more than 320 samples, while StitchNet requires only 32 samples used to stitch the fragments together (realizing over a 90\% reduction). 

Therefore, only a small amount of training samples and computation are required for StitchNet to achieve comparable accuracy. This demonstrates that StitchNet effectively reuses the information already captured in the fragments to bootstrap network creation. This allows for personalization of tasks and on-the-fly training without substantial data requirements.

\subsection{StitchNet on New Tasks}
\label{subsec:differenttasks}

\begin{figure}[th!]
     \centering
     \begin{minipage}[t]{\linewidth}
        %  \centering
    \includegraphics[width=\linewidth]{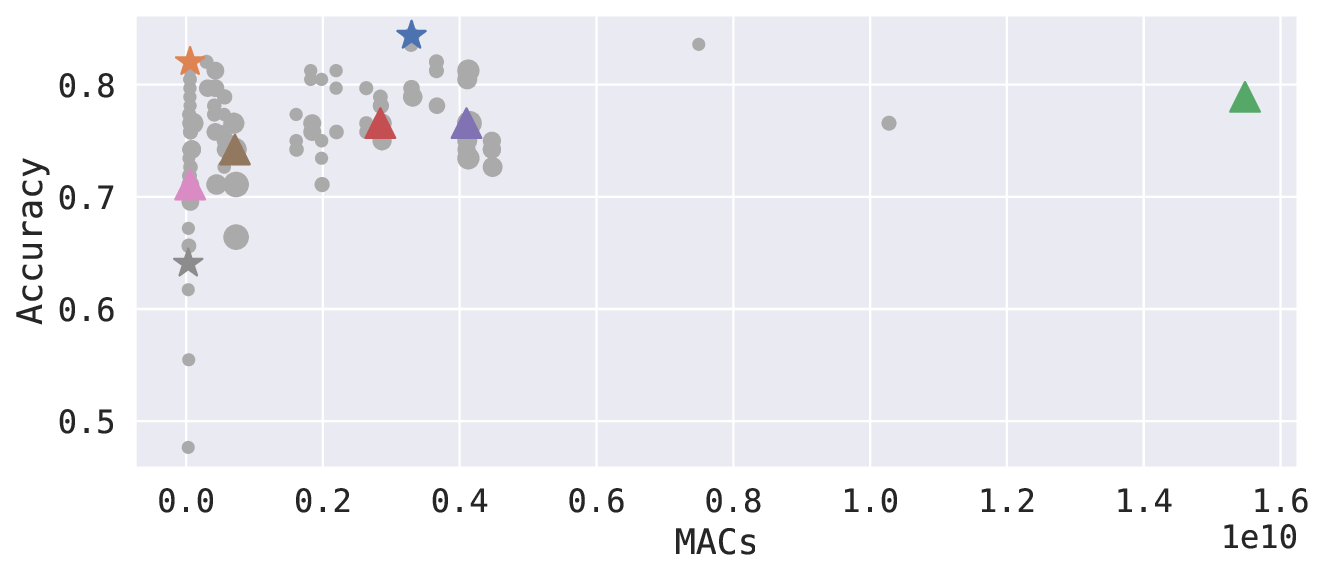}
        % \subcaption{Accuracy vs MACs}
        % \label{fig:accvsmacs_bean}
     \end{minipage}
     % \vfill
     \begin{minipage}[t]{\linewidth}
        \vspace{-12px}
        \centering
        \includegraphics[width=.8\linewidth]{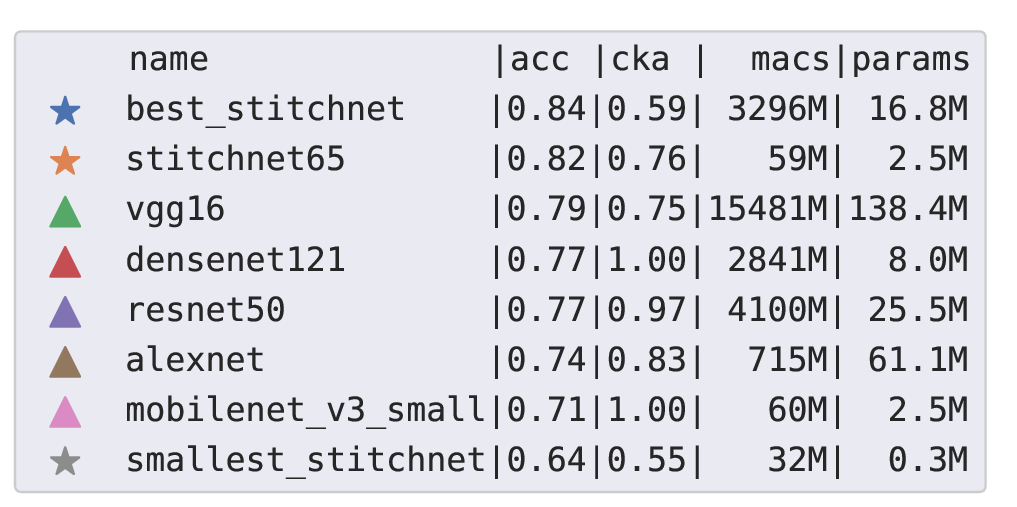}
        % \subcaption{Accuracy vs MACs}
        % \label{fig:accvsmacs_bean_legend}
     \end{minipage}
    \caption{
    Accuracy vs MACs on ``Beans.'' The best StitchNet ($acc$=0.84) with only 16.8M parameters and 3296M MACs performs better than the best pre-trained model(s) (vgg16 with $acc$=0.79 and 15481M MACs).
    }
    \label{fig:accvsmacs_bean}
\end{figure}

We now show that the StitchNet approach of leveraging existing pre-trained network fragments is highly general. It achieves high performance on new specific tasks and domains not explicitly used in training the pre-trained network fragments. 
% It achieves high performance on new tasks and domains that were not part of the training of these pre-trained network fragments. 
To demonstrate this, we use an entirely different dataset for evaluation that shares no overlap with the ImageNet-1K dataset used to train the pre-trained network fragments. 
% Further, to show the sustained computational efficiencies achieved by StitchNet, we do not confine the problem to a subset of classes used for the network training. 
For this experiment, we use the ``Beans'' dataset~\cite{beansdata}, which includes images of diseased and healthy beans leaves. This dataset is divided into three categories: \textit{angular\_leaf\_spot}, \textit{bean\_rust}, and \textit{healthy}. We use 32 samples from the validation split of the dataset to generate StitchNets. Then, we use all the validation split of 133 samples to train a KNN classifier that uses the final 1,000-dimensional vectors obtained from the networks as feature inputs. To evaluate network accuracy, we use the 128-sample test split. The results are shown in Fig.~\ref{fig:accvsmacs_bean}. We see that the top-performing StitchNet with an accuracy ($acc$) of 0.84 and 3296M MACs outperforms the original pre-trained model (vgg16) in both accuracy and efficiency, 0.79 and 15481M MACs respectively.  This shows again that with StitchNet, we can generate the most efficient and smallest neural network tailor-made for this specific task, eliminating the need for retraining the neural networks for different tasks.

% \subsection{On-the-fly Network Selection}

% In this section, we explore the ability to select network to use on the fly based on the CKA score of the new data batch vs the previous data batch.

\subsection{On-the-fly Network Creation and Inference}
\label{subsec:onthefly}

We now delve into a novel set of applications and scenarios that are made possible by StitchNet: on-the-fly neural network creation and inference.  In this scenario, we use a batch of unlabeled images that we wish to use for a particular task (e.g., classification or detection) as our target dataset in the generation of StitchNets. With only a minor modification to the StitchNet algorithm to additionally return task results, the StitchNet generation process can return the inference outputs along with the generated StitchNets.

Let us illustrate how this could be applied in practice. Imagine a world where fragments of pre-trained neural networks for different tasks are indexed and distributed on the Internet. Compatible fragments can be found and organized into a pool from which candidate fragments can be selected and quickly composed to form a new StitchNet for a certain task.  Now, imagine that we want to create a neural network for classifying local food images with only a limited number of labeled images.

Without StitchNet, we either need to train a network from scratch (which may fail due to our limited amount of training data), or find an existing pre-trained neural network, label the dataset, and finetune the network. If the existing pre-trained network is too big or too slow for our use, we will then have to train a new one from scratch. But, with limited amount of data, this task seems impossible.

With StitchNet, we can instead generate a set of candidate StitchNets on-the-fly with the small batch of unlabeled local food images. These StitchNets are created from the pool of existing neural network fragments that have been indexed and distributed over the Internet. The proper fragments can be identified with a given search criteria (e.g., the depth of the network should be less than 5 for computational efficiency reasons, etc.). 
% With only a few labels, we can then use a KNN classifier on top of these output feature vectors to classify new local food images. 
With only the relatively limited computation needed to find stitches rather than the normal high-cost network training methods, we can generate various StitchNets capable of detecting and classifying local food images at the desired computation requirements.

\begin{figure}[th!]
     \centering
     \begin{minipage}[t]{\linewidth}
        %  \centering
    \includegraphics[width=\linewidth]{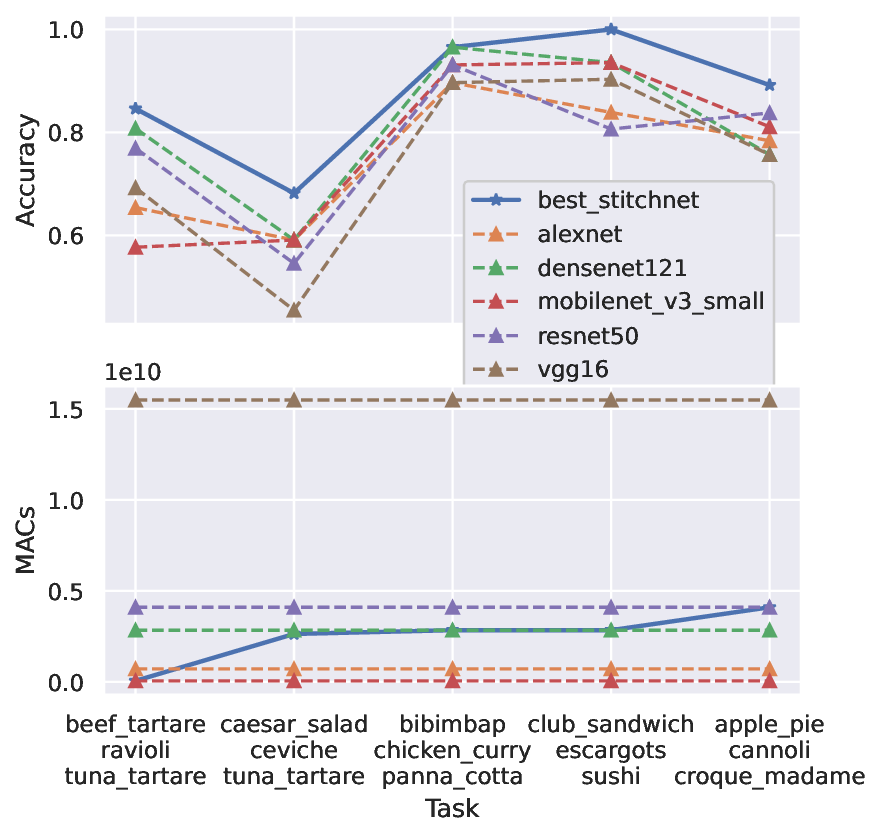}
        % \subcaption{Accuracy vs MACs}
        \label{fig:accvstasks}
     \end{minipage}
     % \vfill
     % \begin{minipage}[t]{\linewidth}
     %   \vspace{-10px}
     %     \centering
     %    \includegraphics[width=\linewidth]{figures2/macsvstasks.eps}
     %    % \subcaption{Accuracy vs MACs}
     %    \label{fig:macsvstasks}
     % \end{minipage}
    \caption{
    Accuracy and MACs as a function of various tasks derived from the ``Food101'' dataset.
    }
    \label{fig:onthefly}
\end{figure}

To demonstrate this new paradigm, we use the ``Food101'' dataset~\cite{bossard14} and method described in Sec.~\ref{subsec:differenttasks} to demonstrate the feasibility of this use case. In this experiment, we explore 5 different classification tasks. Each task consists of 3 classes randomly picked out of 101 classes from the dataset. The classes for each task are listed in the x-axis of Fig.~\ref{fig:onthefly}. Each task has around 300 samples for training and 30 samples for testing. 32 samples of the training set are used to generate StitchNets. As shown in Fig.~\ref{fig:onthefly}, in all tasks top-performing StitchNets surpass all original pre-trained networks. This underscores StitchNet's efficacy in generating efficient and high-performing networks instantly in response to new tasks.

\section{Discussion}
\label{sec:discussion}
% We discuss intuitions, complexity, and limits of StitchNet.%, examine its complexity, and discuss its limitations.
% We now discuss the intuition behind StitchNets, introduce new applications they enable, and discuss their limitations.
% We now discuss the intuition behind StitchNets, examine their complexity and relation to related methods, introduce new applications they enable, and discuss their limitations.

\subsection{Why does StitchNet work?}
\label{section:discuss_why}
We first discuss why we can reuse existing fragments of networks to create new neural networks without retraining.  One core reason is that neural networks tend to learn fundamental and universal features.  Studies~\cite{li2015convergent, lu2018shared, morcos2018insights, wang2018towards, lenc2015understanding, kornblith2019similarity, tang2020similarity} have shown that neural networks learn fundamental features, such as edges for different tasks. Since these learned features are fundamental, they should be reusable rather than relearned.
The challenge, however, is that although these features may be universal, they may not be compatible with one another ``out of the box.''  Therefore, we require the stitching process introduced in Sec.~\ref{section:stitch} to project the fragments into a compatible space.

% Beyond the reuse of universal features and compatibility transformations, StitchNet is enabled by networks being fundamentally redundant. Due to nonlinear activations and built-in redundancies, networks tolerate certain amounts of error.  As such, fragments need not be perfectly compatible individually to produce a network that in aggregate operates performantly.

Beyond the reuse of universal features and compatibility transformations, StitchNet benefits from the inherent redundancy in networks. This redundancy, arising from nonlinear activations and redundant features, allows networks to tolerate certain levels of error. Consequently, individual fragments do not need to be perfectly compatible to form a network that operates effectively as a whole.

\subsection{Complexity Comparison}

We now compare the complexity of the traditional training process using backpropagation with the StitchNet generation process. 
Traditional training complexity is $O(ndp)$, where $n$ is the number of parameters in the network, $p$ is the number of epochs used to train, and $d$ is the size of the dataset.

The complexity of StitchNet generation is $O(nqm)+O(K^L)+O(nq)$. The first term $O(nqm)$ is the evaluation cost of the target dataset of size $q$ on $m$ networks in the pool, where $q \ll d$ and $n$ is the number of parameters in the network (assuming that the networks have the same number of parameters). The second term $K^L$ is the search cost, where $K$ is the span value we search at each level and $L$ is the maximum depth of search. Using a high threshold cutoff $T$ on the overall CKA score keeps search cost $K^L$ small. The third term $O(nq)$ is the cost of the stitching (See Sec.~\ref{section:stitch}). Therefore, for a reasonable setting of hyperparameters ($K,T,L$) in Algorithm~\ref{alg:stitchnet}, StitchNet realizes substantial computation gains over traditional training methods since $q \ll d$ and $m \ll p$.

\subsection{Limitations}
\label{subsec:limitations}

% One limitation is that the target task needs to be a subset (or a composition) of the terminating fragment tasks in the fragment pool. 
While a large pool of network fragments can lead to higher applicability and quality of the generated StitchNets, it can also result in high search costs. Indexing large quantities of networks to form the fragment pool will necessitate the development of novel search methods. We see this as analogous to indexing web pages on the World Wide Web, suggesting a ``Google for Fragments.'' Much like the web search needed to index written content, large amounts of neural network ``content'' need to be indexed for their value to be unlocked.  Early indexing efforts can tag fragments based on dataset characteristics, computational characteristics, etc. More advanced efforts can look at inward and outward connections of each fragment to determine its rank in results. 
Once a narrowed set of fragments is coarsely identified, the efficient procedure introduced in this paper can generate the StitchNets.  
Future work will address indexing and distribution that will enable StitchNet to operate at scale.

\section{Related Work}

Current methods to adapt existing networks to target tasks are transfer learning, 
fine-tuning \cite{weiss2016survey}, 
% fine-tuning \cite{pan2009survey, weiss2016survey}, 
distillation \cite{hinton2015distilling} and neural architecture search (NAS) \cite{pham2018efficient}. A related concept is unsupervised domain adaptation~\cite{wang2018deep,zhang2018collaborative, tzeng2014deep,kumar2018co,shu2018dirt,ben2010theory,saito2017asymmetric}, where an existing network is modified using an unlabeled target dataset. StitchNet works similarly by stitching fragments using an unlabeled target dataset to create a neural network for the target task. What distinguishes StitchNet from most of the previous works is that it eliminates the need for retraining the network.

StitchNet takes advantage of the assumption that fragments have shareable representations. This assumption helps explain why fragments can be stitched together into a coherent high-performing network: dissimilar yet complementary fragments once projected into a similar space are compatible with one another. Several existing works including~\cite{li2015convergent, mehrer2018beware, lu2018shared, morcos2018insights, wang2018towards, lenc2015understanding, kornblith2019similarity, tang2020similarity} have studied this shareable representation assumption.

Gygli, et al.~\cite{gygli2021towards} reuse network components by training networks to produce compatible features by adding regularization during training to make networks compatible. StitchNet, however, focuses on creating neural networks without retraining the network. It is therefore more generally applicable. Several works~\cite{lenc2015understanding, csiszarik2021similarity, bansal2021revisiting, yang2022deep, pan2023stitchable} combine network components by adding a stitching layer and training the recombined network with supervised loss for several epochs. StitchNet differs by approaching the topic through an applied lens focusing on minimizing training costs by adding a parameter-less stitching mechanism, and therefore does not require any retraining. Instead, weights are adapted to be compatible (See~Sec.~\ref{section:stitch}).

% paragraph about NAS

% In parallel, model distillation has emerged as another influential concept in deep learning \cite{hinton2015distilling, lopez2015unifying}. This technique aims to simplify a complex model by transferring its knowledge to a smaller model, thereby enhancing the efficiency of deployment, especially in resource-constrained environments. Although the primary principle of StitchNets is to create neural networks without training, the intersection of StitchNets and network compression, particularly knowledge distillation, is intriguing. By merging these concepts, it might be possible to create a simplified yet effective model that encapsulates the knowledge of the stitched networks. This approach could offer additional performance benefits and flexibility, particularly in scenarios where computational resources are constrained.

\section{Conclusion}
StitchNet is a new paradigm that can take advantage of a growing global library of neural networks to fundamentally change the way neural networks are created.  By reusing fragments of these networks to efficiently compose new neural networks for a given task, StitchNet addresses two of the most fundamental issues that limit the creation and use of neural networks: large data and computation requirements.

StitchNet does this by leveraging CKA as a compatibility measure that guides the selection of network fragments, tailored to specific accuracy needs and computing resource constraints. Our work has shown that neural networks can be efficiently created from compatible network fragments of different models at a fraction of computing resources and data requirements while achieving comparable accuracy. We also explore a novel on-the-fly efficient neural network creation and inference application unlocked by StitchNet. Future work will extend StitchNet to LLMs and additional applications.

\section*{Acknowledgment}
This research was supported in part by the Air Force Research Laboratory under contract number FA8750-22-1-0500 and in part by a generous gift to the Center for Research on Computation and Society at Harvard University in support of research on applied cryptography and society.
% \textcolor{red}{The authors would like to thank [] for their financial support under grant number [].}

\bibliographystyle{IEEEtran}
\bibliography{main.bib}

\end{document}